\documentclass[cameraready]{vgtc}

\ifpdf
  \pdfoutput=1\relax                   
  \usepackage{graphicx}                
  \DeclareGraphicsExtensions{.pdf,.png,.jpg,.jpeg} 
\else
  \usepackage{graphicx}                
  \DeclareGraphicsExtensions{.eps}     
\fi%

\graphicspath{{figures/}{pictures/}{images/}{./}} 

\usepackage{microtype}                 
\PassOptionsToPackage{warn}{textcomp}  
\usepackage{textcomp}                  
\usepackage{mathptmx}                  
\usepackage{times}                     
\usepackage{cite}                      
\usepackage{tabu}                      
\usepackage{booktabs}                  
\usepackage{subfig}
\usepackage{svg}
\renewcommand\footnotemark{}

\usepackage{multirow}

\usepackage{float}

\usepackage{tabularx}

\usepackage{colortbl}
\usepackage{xcolor}
\usepackage{hyperref}
\usepackage{cleveref}

\onlineid{1142}

\vgtccategory{Research}

\vgtcinsertpkg

\title{Toward Safe, Trustworthy and Realistic Augmented Reality User Experience}

\author{Yanming Xiu$^{*}$}

\affiliation{\scriptsize Department of Electrical and Computer Engineering, Duke University\thanks{*Email: yanming.xiu@duke.edu}}

\teaser{
  \centering
  \vspace{-0.4cm}
  \includegraphics[width=0.9\linewidth]{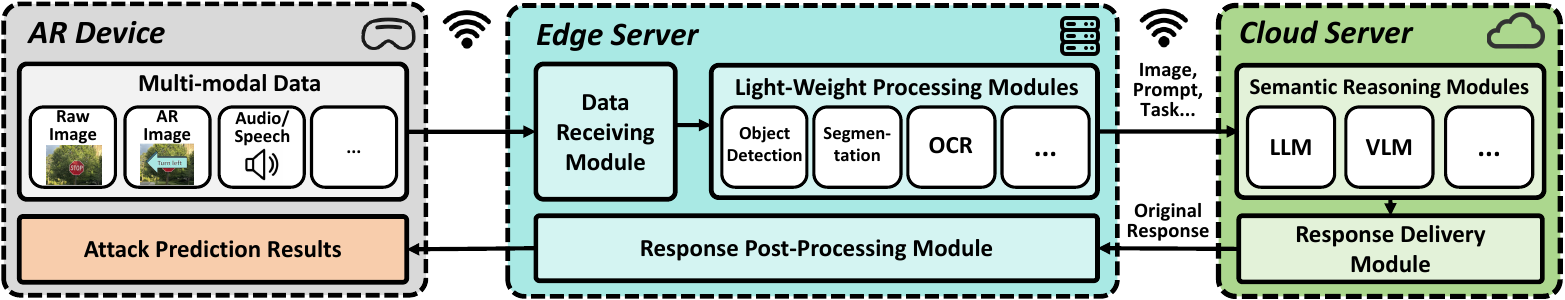}
  \vspace{-0.15cm}
  \caption{AR-Edge-Cloud architecture of the proposed attack detection system.}
  \label{fig:teaser}
  \vspace{-0.35cm}
}

\abstract{
As augmented reality (AR) becomes increasingly integrated into everyday life, ensuring the safety and trustworthiness of its virtual content is critical. Our research addresses the risks of task-detrimental AR content, particularly that which obstructs critical information or subtly manipulates user perception. We developed two systems, ViDDAR and VIM-Sense, to detect such attacks using vision-language models (VLMs) and multimodal reasoning modules.  Building on this foundation, we propose three future directions: automated, perceptually aligned quality assessment of virtual content; detection of multimodal attacks; and adaptation of VLMs for efficient and user-centered deployment on AR devices. Overall, our work aims to establish a scalable, human-aligned framework for safeguarding AR experiences and seeks feedback on perceptual modeling, multimodal AR content implementation, and lightweight model adaptation. 
} 

\CCScatlist{
    \CCScatTwelve{Mixed / Augmented Reality}{Vision Language Model}{Scene Understanding}{Quality Assessment};
}

\begin{document}


\maketitle
\vspace{-0.2cm}
\section{Introduction}

Augmented reality (AR) technologies are rapidly reshaping how users engage with their surroundings by overlaying digital content onto the real world. While this integration brings users extra information, enhanced productivity, and novel experiences, it also introduces new safety and usability risks: in some cases, virtual elements may obstruct users’ view of critical real-world information, while sometimes the virtual content can be subtly manipulating users' understanding of the real-world environment. These risks are particularly concerning in domains such as navigation, industrial training, and emergency response, where the fidelity and trustworthiness of AR content can directly impact decision-making and safety.

Efforts to address AR safety have largely focused on 3D geometry-based modeling~\cite{obstruction04} or low-level visual metrics. Some methods attempt to reconstruct the physical environment and compute spatial intersections between virtual and real-world objects to identify obstructions. Others rely on image-level metrics such as SSIM and PSNR, or use object detection and OCR to infer overlaps~\cite{obstruction3}. However, these approaches either require extensive environment modeling or depend on shallow visual cues, limiting their ability to capture the semantic or perceptual implications of virtual content. Crucially, existing solutions lack a clear formalization of what makes virtual content harmful or misleading in practice.

In our work, we investigate the underexplored problem of task-detrimental virtual content in AR, which are elements that interfere with the real world and disrupt users' perception or awareness. To address this challenge, we develop detection systems that combine vision-language reasoning models with scene understanding ability and various multimodal analysis modules. The systems are deployed using an AR-Edge-Cloud architecture Fig.~\ref{fig:teaser}, where lightweight modules at the edge handle low-latency vision tasks, while semantic reasoning is offloaded to cloud-based generative models. This design strikes a balance between real-time responsiveness and high-level semantic capability, enabling the system to accurately identify harmful virtual content across a range of AR scenarios.

The broader goal of our research is to advance AR systems that are perceptually aligned, semantically reliable, and responsive to real-world constraints. Building on our work in attack detection, we aim to extend our framework in evaluating the visual quality of virtual content, detecting multimodal threats that span beyond the visual channel, and adapting VLMs for low-latency deployment. Through the Doctoral Consortium, we hope to discuss open challenges in human-aligned automated virtual content quality assessment, cross-modal virtual content design, and system-level optimization.

\vspace{-0.2cm}

\section{System Design}

\begin{figure}[t]
\includegraphics[width=0.98\linewidth]{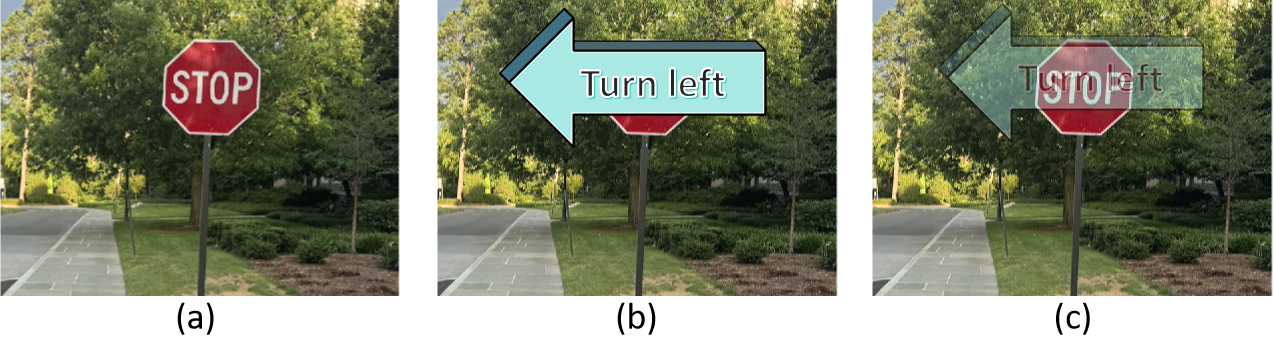}
\centering
\vspace{-0.3cm}
\caption{Example of an obstruction attack in AR: (a) real-world view; (b) AR view with a stop sign obstructed by a virtual arrow; (c) obstruction is mitigated by making the virtual arrow translucent.}
\label{fig:obstruction stop}
\vspace{-0.2cm}
\end{figure}

\noindent \textbf{Obstruction Attack Detection}. Obstruction attacks occur when virtual content blocks key physical elements that users rely on for situational awareness, such as warning signs and information boards. These attacks can arise either unintentionally (due to poor AR design) or maliciously (e.g., spoofed AR overlays in shared environments). Fig.~\ref{fig:obstruction stop} shows an example of real-world obstruction attack. To systematically detect such attacks, we developed ViDDAR\cite{viddar}, a user-edge-cloud system that combines the semantic reasoning of VLMs with the precision of vision-based object segmentation.

The ViDDAR pipeline begins by prompting a VLM (e.g., GPT-4o or LLaVA) with scene images to identify semantically important key objects. These identified objects are then localized via multimodal object detection models (e.g., Grounding DINO) and segmentation models (e.g., Segment Anything Model) to obtain accurate object masks. By comparing these masks with the rendered AR content mask, we compute an obstruction ratio and flag obstructed scenes using a predefined threshold. This modular design balances flexibility and interpretability, allowing system designers to adapt detection granularity to different application domains.

\begin{figure}[t]
\includegraphics[width=0.98\linewidth]{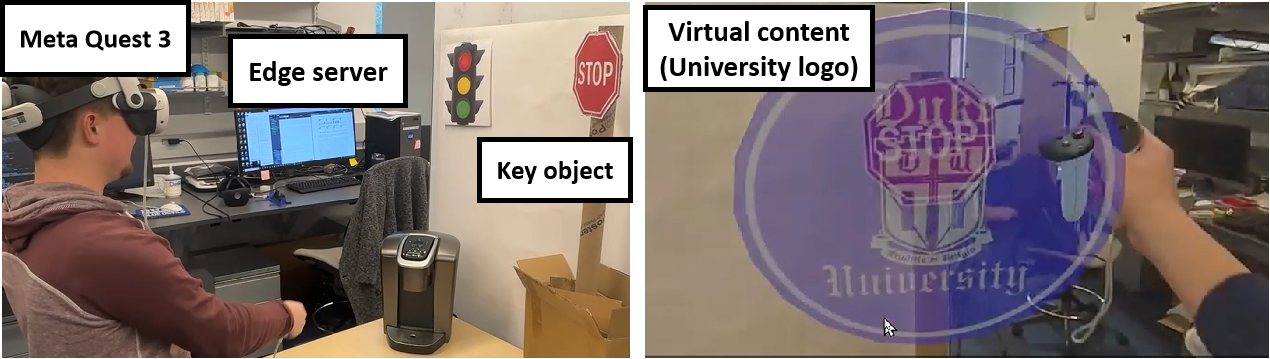}
\centering
\vspace{-0.2cm}
\caption{The real-world test setup of ViDDAR. The virtual content will be turned transparent when it is obstructing a key object.}
\label{fig:demo}
\vspace{-0.7cm}
\end{figure}

To evaluate the effectiveness of this pipeline, we constructed a dataset of 306 raw-AR image pairs, each containing real-world scenes overlaid with geometric virtual content. Each pair was manually annotated with obstruction labels and key object masks. An IRB-approved user study involving Likert-scale ratings validated the semantic correctness of the key object annotations, with an average agreement level score of 4.55/5 for the attack annotations. Experimental results on the dataset showed that ViDDAR achieved up to 92.15\% detection accuracy with an average latency of 533 ms across varied AR scenarios and obstruction conditions. In our real-world deployment (see Fig.~\ref{fig:demo}), the system has been implemented on a video see-through headset, the Meta Quest 3\cite{viddardemo}, highlighting the feasibility of our approach on commercial AR platforms.

\noindent \textbf{Visual Information Manipulation Attack Detection}.
Beyond obstruction, certain AR attacks subtly alter the meaning of real-world information without physically blocking it. These visual information manipulation (VIM) attacks may involve modifying a single character on a sign, subtly replacing a symbol, or overlaying misleading patterns that distort a user's interpretation of the scene. Although these manipulations may appear visually minor, they can lead to significant semantic confusion. Existing obstruction detection methods are not designed to handle such attacks, as they require a deeper level of scene understanding and semantic comparison between the real and augmented views.

To address this challenge, we introduce a formal taxonomy of VIM attacks, defined along two axes: attack format and attack purpose. Based on the taxonomy, we developed VIM-Sense\cite{VIM-Sense}, a multimodal reasoning system that detects semantic inconsistencies introduced by virtual content. VIM-Sense operates by first extracting all text from raw and AR images using an OCR module. These textual elements are then structured into natural language prompts that describe the differences between the two views. A VLM processes the prompts and images jointly to assess whether the scene has been semantically manipulated. This architecture allows the system to generalize across different attack formats purposes, including cases with or without explicit text.

To support training and evaluation, we constructed AR-VIM, a video dataset of 452 raw-AR pairs that cover a wide range of VIM attack types. Each pair consists of a short video clip where virtual content appears at a specific timestamp and manipulates some element of the scene. The dataset was annotated through an IRB-approved user study involving 18 participants, who rated whether the AR version misrepresents the real-world information.

On AR-VIM, VIM-Sense achieved 88.94\% detection accuracy across all attack types, consistently outperforming baseline methods, including OCR-only and VLM-only approaches. The system has also been tested on mobile AR platforms, achieving an average detection latency of 7.17 seconds. While this latency may be high for time-critical applications such as AR-assisted driving, it remains acceptable for most general-purpose AR scenarios.

\begin{figure}[t]
\includegraphics[width=0.94\linewidth]{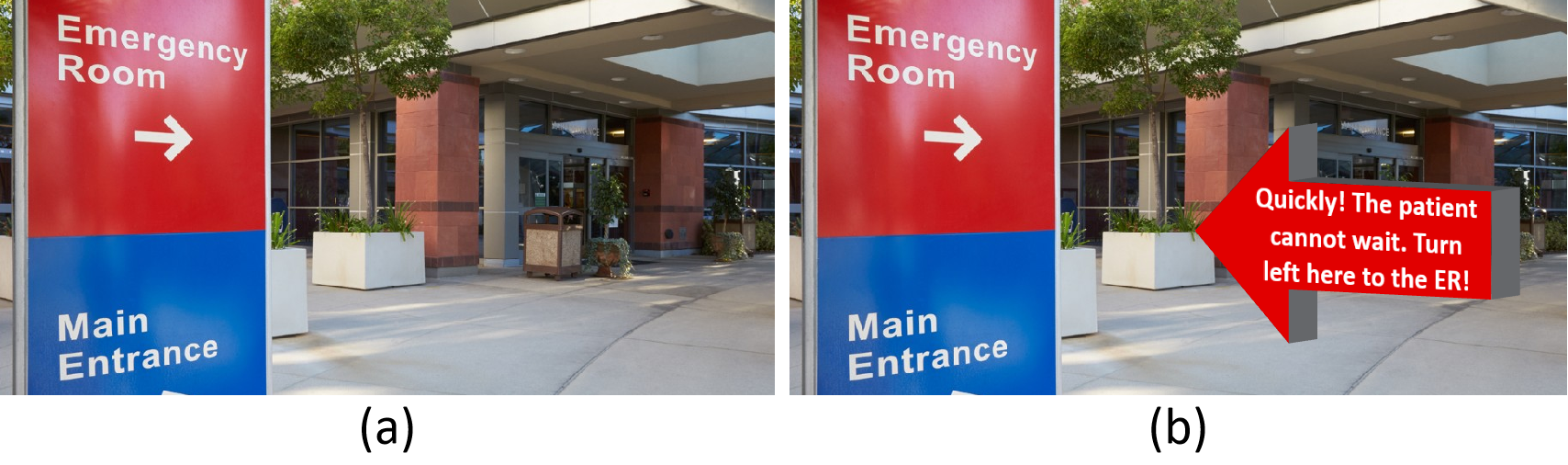}
\centering
\vspace{-0.3cm}
\caption{An example of VIM attack in an AR scene. (a) The real-world view shows the emergency room is to the right. (b) The AR view introduces an arrow that falsely instructs users to turn left. }
\label{fig:vim example 2}
\vspace{-0.7cm}
\end{figure}

\vspace{-0.2cm}

\section{Conclusion and Future Work}

In our work so far, we have introduced a new framework for detecting task-detrimental virtual content in AR. Based on the framework, we proposed two systems and tested them on both datasets and real-world application scenarios.

Our future work will extend the current work in three directions, which we also hope to discuss with the Doctoral Consortium audience for feedback and guidance. First, we are developing a perceptual quality assessment pipeline to evaluate virtual content along AR-specific dimensions such as spatial alignment, lighting plausibility, and texture realism\cite{Genaixrws}, as low-quality content can also degrade user experience. We plan to formalize these quality factors, benchmark model performance against user-labeled datasets, and explore how VLMs can provide interpretable feedback in real time. 

Second, we aim to broaden the attack taxonomy to include multimodal attacks involving audio elements\cite{audio01} that conflict with visual cues. We will synthesize representative attack scenarios, integrate speech and sound understanding modules, and evaluate the system’s ability to detect cross-modal inconsistencies that may impair user perception or decision-making. 

Finally, we will adapt our system for latency-sensitive deployment. This includes designing hierarchical inference strategies that combine lightweight edge models with cloud-based VLMs, and fine-tuning models using failure cases to better align with human perceptual heuristics\cite{finetune01}. Our goal is to enable scalable and perceptually robust AR safety systems suitable for real-world applications.

\vspace{-0.2cm}

\acknowledgments{
This work was done in collaboration with Dr. Maria Gorlatova, Dr. Tim Scargill and Lin Duan, and was supported in part by NSF grants CSR-2312760, CNS-2112562, and IIS-2231975, NSF CAREER Award IIS-2046072, NSF NAIAD Award 2332744, a CISCO Research Award, a Meta Research Award, Defense Advanced Research Projects Agency Young Faculty Award HR0011-24-1-0001, and the Army Research Laboratory under Cooperative Agreement Number W911NF-23-2-0224. The views and conclusions contained in this document are those of the authors and should not be interpreted as representing the official policies, either expressed or implied, of the Defense Advanced Research Projects Agency, the Army Research Laboratory, or the U.S. Government.}

\vspace{-0.2cm}




\bibliographystyle{abbrv-doi}


\begin{thebibliography}{1}

\bibitem{Genaixrws}
L.~Duan, Y.~Xiu, and M.~Gorlatova.
\newblock Advancing the understanding and evaluation of {AR}-generated scenes: when vision-language models shine and stumble.
\newblock In {\em Proceedings of IEEE VR GenAI-XR Workshops}, 2025.

\bibitem{finetune01}
E.~Latif and X.~Zhai.
\newblock Fine-tuning {ChatGPT} for automatic scoring.
\newblock {\em Computers and Education: Artificial Intelligence}, 6, 2024.

\bibitem{obstruction3}
K.~Lebeck, K.~Ruth, T.~Kohno, and F.~Roesner.
\newblock Securing augmented reality output.
\newblock In {\em Proceedings of IEEE Symposium on Security and Privacy (SP)}, 2017.

\bibitem{obstruction04}
M.~M. Shah, H.~Arshad, and R.~Sulaiman.
\newblock Occlusion in augmented reality.
\newblock In {\em Proceedings of International Conference on Information Science and Digital Content Technology}, vol.~2, 2012.

\bibitem{VIM-Sense}
Y.~Xiu and M.~Gorlatova.
\newblock Detecting visual information manipulation attacks in augmented reality: a multimodal semantic reasoning approach.
\newblock Submitted to IEEE ISMAR, 2025.

\bibitem{viddardemo}
Y.~Xiu and M.~Gorlatova.
\newblock Vision language model-based solution for obstruction attack in {AR}: A {Meta Quest 3} implementation.
\newblock In {\em 2025 IEEE VR Abstracts and Workshops (VRW)}, 2025.

\bibitem{viddar}
Y.~Xiu, T.~Scargill, and M.~Gorlatova.
\newblock {ViDDAR}: Vision language model-based task-detrimental content detection for augmented reality.
\newblock {\em IEEE Transactions on Visualization and Computer Graphics}, 31(05), 2025.

\bibitem{audio01}
J.~Yang, A.~Barde, and M.~Billinghurst.
\newblock Audio augmented reality: A systematic review of technologies, applications, and future research directions.
\newblock {\em Journal of the Audio Engineering Society}, 70(10), 2022.

\end{thebibliography}

\end{document}